\documentclass[preprint,authoryear,12pt]{elsarticle}

\graphicspath{{figures/}}
\usepackage[letterpaper,margin=1in]{geometry}
\usepackage{amsmath}
\usepackage{amssymb}
\usepackage{booktabs}
\usepackage[section]{placeins}
\usepackage[hidelinks]{hyperref}

\journal{Transportation Research Part C: Emerging Technologies}

\begin{document}

\begin{frontmatter}

\title{VLM-CASE: vision-language model enabled context-adaptive safety envelopes for anticipatory safe autonomous driving}

\author[psu]{Tianjia Yang}
\ead{tjyang@psu.edu}
\author[sbu]{Ke Li}
\ead{ke.li.1@stonybrook.edu}
\author[sbu]{Ruwen Qin}
\ead{Ruwen.Qin@stonybrook.edu}
\author[psu]{Xianbiao (XB) Hu\corref{cor1}}
\ead{xbhu@psu.edu}
\cortext[cor1]{Corresponding author.}

\affiliation[psu]{organization={Department of Civil and Environmental Engineering, The Pennsylvania State University},
  city={University Park}, state={PA}, postcode={16802}, country={United States}}
\affiliation[sbu]{organization={Department of Civil Engineering, Stony Brook University},
  city={Stony Brook}, state={NY}, postcode={11794}, country={United States}}

\begin{abstract}
Adverse driving conditions, such as bad weather, remain a principal barrier to autonomous driving because they degrade two things at once: what the vehicle can perceive and what it can physically do. Human drivers cope by anticipation, reasoning about the scene and re-budgeting speed, following distance, and steering before grip or sight is lost, whereas current autonomous driving systems at best react after the fact. This paper proposes VLM-CASE, a framework that gives an autonomous vehicle this anticipatory capacity while keeping its motion bounded by a formal safety model at all times. A vision-language model (VLM), fine-tuned with low-rank adaptation (LoRA), reasons about the scene from the front-camera image and reports the road surface and visibility conditions. This output parametrizes a context-adaptive safety envelope (CASE), derived from physical limits and the guarantees of responsibility-sensitive safety, that couples braking and steering through a shared friction budget. A model predictive controller then drives freely within the envelope, while the VLM runs asynchronously so it never blocks the real-time control loop. We validate the framework in closed-loop CARLA simulation on tasks that demand both lateral and longitudinal control, across a range of weather, road-surface, and lighting conditions. The resulting controller, VLM-CASE-MPC, completes all trials, outperforming a conventional MPC baseline and a state-of-the-art VLM-integrated controller. Ablations confirm that the gains come from context adaptation, with the friction and visibility adaptations proving complementary. Furthermore, the framework is controller-agnostic and pairs with almost any low-level controller, offering a promising direction for safe autonomous driving. The dataset and supplementary materials for VLM-CASE are available at \url{https://github.com/ytj254/VLM-CASE}.

\end{abstract}


\end{frontmatter}

\section{Introduction}
\label{sec:introduction}

Human drivers handle adverse conditions by anticipation; current autonomous driving systems, at best, by reaction. A driver who notices snow slows before the wheels lose grip, and drops back from the car ahead as fog sets in. Instead of measuring friction and visibility, or consulting a forecast, the driver \emph{reasons} about the scene ahead and adjusts the driving behavior (speed, following distance, steering) to match what the vehicle can do and what the driver can see. Autonomous driving systems lack this anticipatory capacity, making adverse driving conditions one of the principal barriers to their deployment \citep{neumeisterAutomatedVehiclesAdverse2019, boyapatiAutomatedVehiclesAdverse2023}. Adverse conditions degrade driving in two ways. First, they weaken perception by reducing visibility and impairing lane and object detection \citep{zangImpactAdverseWeather2019, zhangPerceptionSensingAutonomous2023}. Second, they alter vehicle maneuverability by changing tire--road friction and therefore its physical capability \citep{wangTireRoadFriction2022}. Conditions such as rain and snow impair perception and maneuverability at once, and the effects are coupled: the vehicle must make safety-critical decisions precisely when both its environmental awareness and its physical capability are reduced.

Existing autonomous driving systems handle adverse conditions in ways that fall short of the human pattern. Perception research under adverse conditions restores what the vehicle can see through robust sensing and fusion \citep{zhangPerceptionSensingAutonomous2023}, yet recovers sensing capability without translating the remaining visibility loss into adapted driving behavior. Friction estimation gauges what the vehicle can do, but only once the surface is already underneath the wheels, too late for an anticipatory response \citep{khaleghianTechnicalSurveyTireroad2017, wangTireRoadFriction2022}. Fixed worst-case safety models such as Responsibility-Sensitive Safety (RSS) \citep{shalev-shwartzFormalModelSafe2017} and control barrier functions (CBFs) \citep{amesControlBarrierFunction2017} impose principled constraints, but their parameters are calibrated offline. Parameters conservative enough for ice cripple driving on dry asphalt, while parameters tuned for asphalt turn unsafe on ice \citep{koopmanAutonomousVehiclesMeet2019a}. Vision-language models (VLMs) can supply the scene understanding these responses lack: shown a snow-covered road, a VLM \emph{knows} that snow implies low friction \citep{zhouVisionLanguageModels2024, tianDriveVLMConvergenceAutonomous2025}. Recent work accordingly uses VLMs to tune controller parameters \citep{longVLMMPCModelPredictive2026}, shape learning rewards \citep{huangVlmrlUnifiedVision2025}, or output actions directly through vision-language-action (VLA) models \citep{hwangEMMAEndtoEndMultimodal2024, xuDriveGPT4InterpretableEndtoEnd2024, zhouAutoVLAVisionLanguageActionModel2025}. Yet across all of them, scene understanding flows into the driving decision but never into the vehicle's safety limits: the bounds on what it may do. Those bounds stay fixed regardless of what the model understands, so any added caution carries no safety guarantee.

What is still missing is a formulation that makes the vehicle's safety limits a function of scene understanding, while preserving the formal safety guarantee. Two challenges make this difficult. The first is to connect a foundation model to the controller: the model offers a semantic interpretation of the scene, while the controller acts on a quantitative set of admissible actions backed by a formal guarantee. Joining two such different objects so that the result still carries that guarantee is nontrivial. The second lies in the safety model itself. Adverse conditions do not merely shrink the vehicle's physical limits; they couple braking and steering, since on a low-friction surface the two draw on one shared friction budget. A safety model built on fixed, decoupled worst-case rules is therefore either unsafe on a low-friction curve or needlessly restrictive elsewhere. The challenge is to formulate a safety model that captures these coupled, condition-dependent physical limits while still admitting a formal guarantee that a safe action always remains available.

\begin{figure}[!ht]
  \centering
  \includegraphics[width=\textwidth]{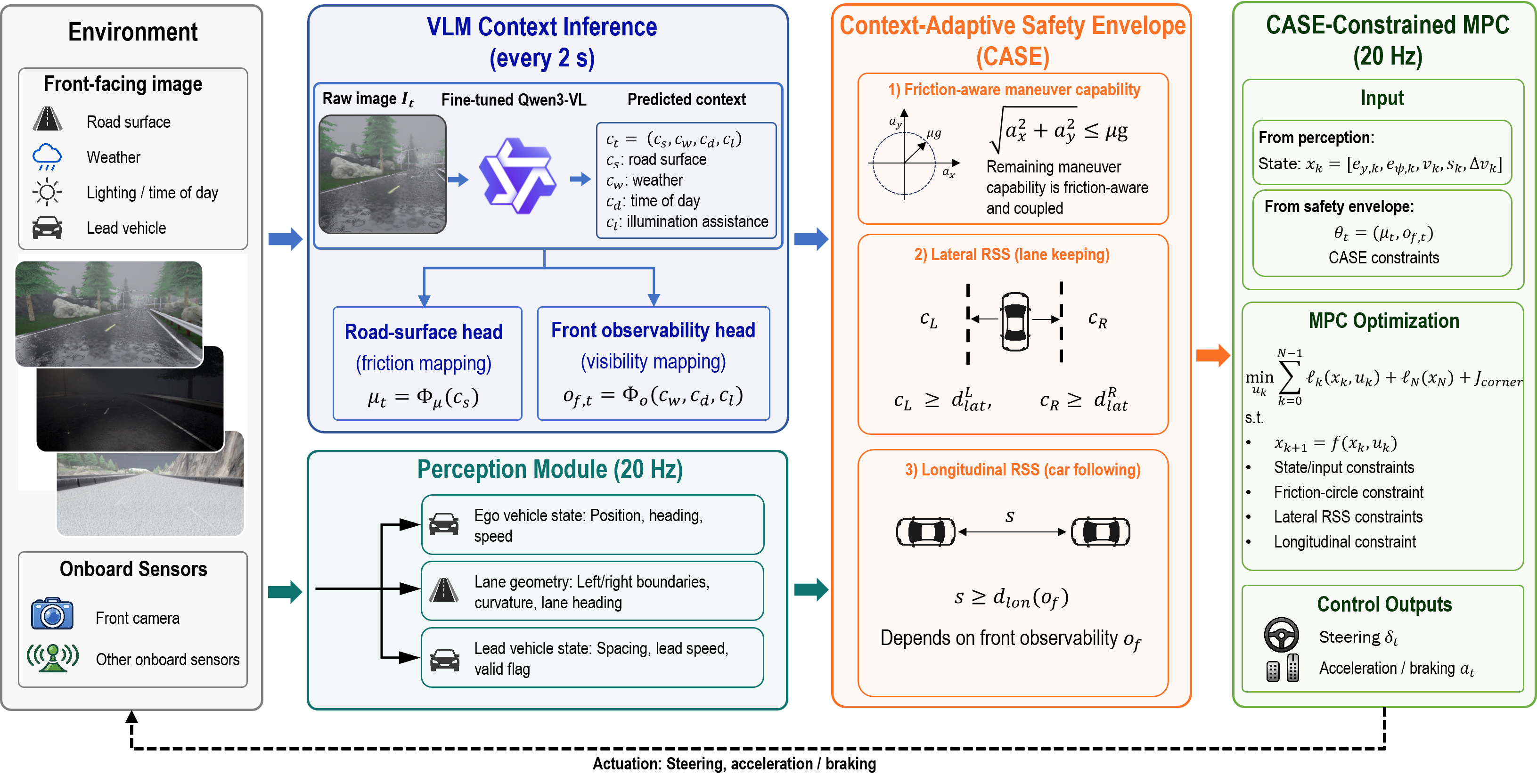}
  \caption{Overview of VLM-CASE, an anticipatory, safety-guaranteed autonomous driving framework.}
  \label{fig:framework}
\end{figure}

We propose VLM-CASE, a framework that resolves both challenges by letting scene understanding set the vehicle's safety limits while capturing how adverse conditions couple them. This marks a shift in how scene understanding reaches control. In the prevailing paradigm, exemplified by VLM-MPC \citep{longVLMMPCModelPredictive2026} and VLM-RL \citep{huangVlmrlUnifiedVision2025}, it flows into the driving decision while the safety limits stay fixed. In VLM-CASE, it flows into the safety envelope, which bounds the vehicle's action with a formal guarantee. As shown in Figure~\ref{fig:framework}, the framework has three components. A VLM, fine-tuned with low-rank adaptation (LoRA), reasons about the driving scene from the front camera and reports the safety-relevant conditions. These conditions parametrize a \emph{context-adaptive safety envelope} (CASE), a quantitative set of admissible actions formulated from physical constraints and the formal guarantees of RSS. Its parameters couple the two effects of adverse conditions in one constraint set: lower friction shrinks the feasible accelerations and ties braking to steering through a shared friction budget, while reduced visibility widens the required following margins. An MPC controller, one realization among many, then drives freely within the envelope, so the vehicle gains the anticipation that scene understanding provides while what it is allowed to do stays formally bounded. The VLM runs asynchronously, so its second-scale inference never blocks the millisecond-scale control loop. We validate VLM-CASE through closed-loop simulation in CARLA across a range of adverse driving conditions.

The contributions of this paper are fourfold:
\begin{itemize}
  \item \textbf{VLM-driven anticipatory safety.} We propose a framework in which the VLM's scene understanding sets the vehicle's formal safety limits instead of only guiding its driving decisions, giving autonomous driving the anticipatory capacity of human drivers while preserving the safety guarantee.
  \item \textbf{Context-adaptive safety envelope.} We realize this framework through the context-adaptive safety envelope, a context-parametrized admissible action set that couples lateral and longitudinal RSS constraints through a shared friction budget.
  \item \textbf{Asynchronous architecture.} We present an asynchronous architecture that integrates second-scale semantic inference with millisecond-scale safety-constrained control without compromising real-time operation.
  \item \textbf{Closed-loop evidence.} We provide closed-loop empirical evidence that anticipatory, human-like caution (slowing before curves on snow, extending headway in rain at night) can substantially enhance autonomous driving safety.
\end{itemize}

The remainder of this paper is organized as follows. Section~\ref{sec:literature} reviews related work. Section~\ref{sec:methodology} formulates the safety envelope and describes the framework. Section~\ref{sec:experiments} details the experimental setup, Section~\ref{sec:results} presents results, and Section~\ref{sec:conclusion} concludes.

\section{Literature review}
\label{sec:literature}

Research related to this work falls into three streams: autonomous driving under changing environmental conditions, safety-constrained vehicle control, and vision-language models for autonomous driving. This section reviews each stream in turn.

\subsection{Autonomous driving under changing environmental conditions}
\label{sec:lit-environment}

Adverse weather challenges many elements of the autonomous driving task: rain, fog, and snow attenuate and scatter the signals on which cameras, lidar, and radar depend, obscure lane markings, and reduce the range at which other road users can be detected \citep{zangImpactAdverseWeather2019, zhangPerceptionSensingAutonomous2023, yangSimLKASSimulationbasedFramework2025}. At the same time, precipitation changes the road surface itself, cutting the available tire--road friction by half or more between dry and snow-covered asphalt. Field evaluations of autonomous vehicle technology in adverse weather consistently identify this dual degradation, of information and of physical capability, as a principal barrier to deployment beyond fair-weather operational design domains \citep{neumeisterAutomatedVehiclesAdverse2019, boyapatiAutomatedVehiclesAdverse2023}.

A first line of response targets perception. Multimodal fusion exploits the complementary strengths of camera, lidar, and radar to keep detection alive in fog and rain \citep{bijelicSeeingFogSeeing2020}, point-cloud de-noising removes the clutter that falling snow introduces into lidar returns \citep{heinzlerCNNBasedLidarPoint2020}, and a growing body of work benchmarks and improves the robustness of detection and segmentation under weather corruptions \citep{songRobustnessAware3DObject2024b, dingClearAdverseSurvey2026a}. Runtime sensor monitoring complements these methods by detecting online how severely the sensors are degraded so that downstream systems can respond \citep{sezginSafeAutonomousDriving2023}. A smaller body of work accepts that perception will remain imperfect and hardens the decision layer instead: robust perception-based control retains safety guarantees under bounded perception errors \citep{deanRobustGuaranteesPerceptionBased2020}, and perception-aware MPC propagates perception uncertainty into the controller's constraints \citep{bonzaniniPerceptionAwareChanceConstrainedModel2021}.

A second line of response targets the vehicle's physical capability, such as the tire--road friction. Dynamics-based estimators infer friction from the vehicle's own response (wheel slip, lateral force buildup, yaw-rate deviation) and are accurate once the tires are sufficiently excited, but they identify the surface the vehicle is already driving on \citep{khaleghianTechnicalSurveyTireroad2017, wangTireRoadFriction2022, liuSurveyComprehensiveTaxonomy2025}. Vision-based methods move the estimate ahead of the vehicle by classifying or regressing surface condition from camera images \citep{zhaoRoadFrictionEstimation2024}, fusion schemes combine the two for faster convergence at surface transitions \citep{zhaoTireRoadFrictionCoefficients2025}, and friction-adaptive controllers embed such estimates into the prediction model or constraints \citep{vaskovFrictionAdaptiveStochasticNonlinear2024, liPathTrackingVaryingvelocity2024}. 

Both lines strengthen individual components, but neither translates an overall understanding of the driving conditions into the safety limits the vehicle respects.

\subsection{Safety-constrained vehicle control}
\label{sec:lit-safety}

Formal safety models constrain vehicle behavior through explicit rules with proven guarantees. RSS formalizes minimum longitudinal and lateral distances from worst-case assumptions about response times and accelerations, such that a vehicle executing the proper response never causes a collision \citep{shalev-shwartzFormalModelSafe2017}. CBFs keep the vehicle inside a designated safe set through a quadratic-program filter on the control input \citep{amesControlBarrierFunction2014, amesControlBarrierFunction2017, heRuleBasedSafetyCriticalControl2021}. Such models encode physical assumptions, such as response times, braking capabilities, and safety margins, as fixed parameters chosen at design time.

In practice, however, no single parameter set is appropriate for all conditions. Calibration studies using naturalistic driving data show that parameters matching average human behavior may be neither conservative enough for adverse conditions nor efficient enough for nominal ones \citep{xuCalibrationEvaluationResponsibilitySensitive2021, liuCalibrationEvaluationResponsibilitysensitive2021, chaiEvaluationOptimizationResponsibilitySensitive2020}. \citet{koopmanAutonomousVehiclesMeet2019a} make the underlying problem explicit: braking capability and road friction vary substantially across operational conditions, so parameters fixed at design time are either unsafe or overly conservative once the vehicle meets the physical world. \citet{salviSafetyImplicationsRuntime2022} further show that adapting RSS parameters preserves safety across changing operating conditions. Accordingly, efforts to move beyond fixed parameters have emerged. Robust variants strengthen the guarantee against disturbances \citep{alanControlBarrierFunctions2023} or calibrate the assumed noise level online from recent vehicle trajectories \citep{qiRobustResponsibilitysensitiveSafety2025}, while \citet{lyuAdaptiveSafeMerging2022} adjust the barrier parameters to the characteristics of surrounding drivers. Related robust and adaptive tracking controllers maintain path-following performance under model uncertainty and disturbances \citep{liRobustAdaptiveLearningBased2022}, though without an explicit environmental knowledge source to inform the adaptation.

These methods enforce safety constraints effectively, but their parameters lack a runtime environmental knowledge source, and the longitudinal and lateral rules are typically treated independently.

\subsection{Vision-language models for autonomous driving}
\label{sec:lit-vlm}

Trained on web-scale image and text corpora, VLMs carry broad world knowledge that task-specific driving models lack. This knowledge lets them understand the driving environment from camera input and reason about what it implies, including rare situations outside any training distribution \citep{zhouVisionLanguageModels2024}. In general, the typical pipeline couples a VLM, as the high-level decision maker, to a low-level controller. DriveVLM pairs a VLM's scene analysis with a conventional planning pipeline in a dual slow--fast architecture \citep{tianDriveVLMConvergenceAutonomous2025}. VLM-MPC has the VLM set desired speed, headway, and cost weights for a longitudinal MPC at low frequency \citep{longVLMMPCModelPredictive2026}. VLM-RL uses language goals to shape rewards for reinforcement learning \citep{huangVlmrlUnifiedVision2025}. In parallel, vision-language-action (VLA) models go further: the model itself becomes the driving policy. Originating in robotic manipulation \citep{kimOpenvlaOpenSourceVisionLanguageAction2024}, the paradigm has moved rapidly into driving \citep{sapkotaVisionLanguageActionModelsConcepts2025, jiangSurveyVisionLanguageActionModels2025}. EMMA maps raw camera input directly to planned trajectories \citep{hwangEMMAEndtoEndMultimodal2024}, and DriveGPT4 produces low-level control signals end to end alongside textual explanations \citep{xuDriveGPT4InterpretableEndtoEnd2024}. AutoVLA unifies chain-of-thought reasoning with physically feasible trajectory tokens in a single autoregressive policy \citep{zhouAutoVLAVisionLanguageActionModel2025}. These models, however, place a large network inside the control loop, and the resulting inference latency remains a practical concern \citep{sapkotaVisionLanguageActionModelsConcepts2025, jiangSurveyVisionLanguageActionModels2025}.

Across these architectures, scene understanding influences the driving decision through planning, parameters, rewards, or actions, yet none of them carries a safety guarantee within the system's structure.

\section{Methodology}
\label{sec:methodology}

This section presents the proposed VLM-CASE framework, shown in Figure~\ref{fig:framework}. We first give a system overview, then detail its three modules in turn: the VLM context inference, the context-adaptive safety envelope, and its MPC realization.

\subsection{System overview}
\label{sec:overview}

The proposed framework aims at safety-guaranteed autonomous driving under varying environmental conditions: the vehicle adapts its behavior to the conditions it encounters, while its motion remains bounded by a formal safety model at all times. We study this goal on combined lane keeping and car following, two basic driving tasks whose safe execution depends directly on the road surface and the visibility. The framework computes the control action through the composition
\begin{equation}
\mathbf{u}_t \;=\; \pi\!\left(\mathbf{x}_t;\; \mathcal{U}_{\mathrm{safe}}\!\left(\mathbf{x}_t,\, \Phi\!\left(\Psi\!\left(I_t;\, P\right)\right)\right)\right),
\label{eq:composition}
\end{equation}
where $I_t$ is the front-camera image and $\mathbf{x}_t$ the measured vehicle state at control time $t$. The composition chains three modules: a VLM context inference $\Psi$, a context-to-parameter mapping $\Phi$, and an MPC policy $\pi$ constrained by a safety envelope $\mathcal{U}_{\mathrm{safe}}$. Each module is specified below.

The VLM, fine-tuned for this task, reasons about the driving scene and reports a discrete semantic context
\begin{equation}
\mathbf{c}_t \;=\; \Psi\!\left(I_t;\, P\right) \in \mathcal{C},
\label{eq:vlm-map}
\end{equation}
where $P$ is a fixed instruction prompt and $\mathcal{C}$ is a finite set of categorical judgments about the road surface and the visibility conditions (Section~\ref{sec:vlm-layer}).

The mapping $\Phi$ converts the semantic context into quantitative parameters
\begin{equation}
\boldsymbol{\theta}_t \;=\; \Phi\!\left(\mathbf{c}_t\right) \;=\; \left[\mu_t,\; o_{f,t}\right] \in \Theta,
\label{eq:context}
\end{equation}
where $\mu \in (0,1]$ is the tire--road friction coefficient and $o_f \in [0,1]$ is the forward observability, how reliably the road ahead and the lead vehicle can be perceived. The two parameters capture how adverse conditions degrade driving: $\mu$ bounds what the vehicle \emph{can do}, while $o_f$ bounds what it \emph{can know}.

The parameters $\boldsymbol{\theta}$ are then passed to the \emph{safety envelope}
\begin{equation}
\mathcal{U}_{\mathrm{safe}}(\mathbf{x}, \boldsymbol{\theta}) \;\subseteq\; \mathcal{U},
\label{eq:envelope-set}
\end{equation}
the set of control actions $\mathbf{u} = [a, \delta] \in \mathcal{U}$, with $a$ the commanded longitudinal acceleration and $\delta$ the front steering angle, that remain safe at state $\mathbf{x}$ given the parameters $\boldsymbol{\theta}$, formulated from physical constraints and the formal safety guarantees of RSS (Section~\ref{sec:rss-layer}). The MPC policy $\pi$ then optimizes driving performance over a prediction horizon subject to the envelope at every step and applies the first optimized action (Section~\ref{sec:mpc-layer}).

The composition in Eq.~\eqref{eq:composition} makes this explicit: scene understanding reaches the vehicle's motion through $\boldsymbol{\theta}$, a compact and interpretable set of environment parameters, which set the boundaries of the admissible action set, the safety envelope. All driving decisions are then made by the controller inside the envelope. The VLM and the controller run on different timescales: the VLM performs semantic inference every 2 seconds, while the controller runs at 20~Hz, so the slow reasoning never blocks the real-time control loop (Section~\ref{sec:vlm-layer}).

\subsection{VLM context inference}
\label{sec:vlm-layer}

\begin{figure}[!ht]
  \centering
  \includegraphics[width=\textwidth]{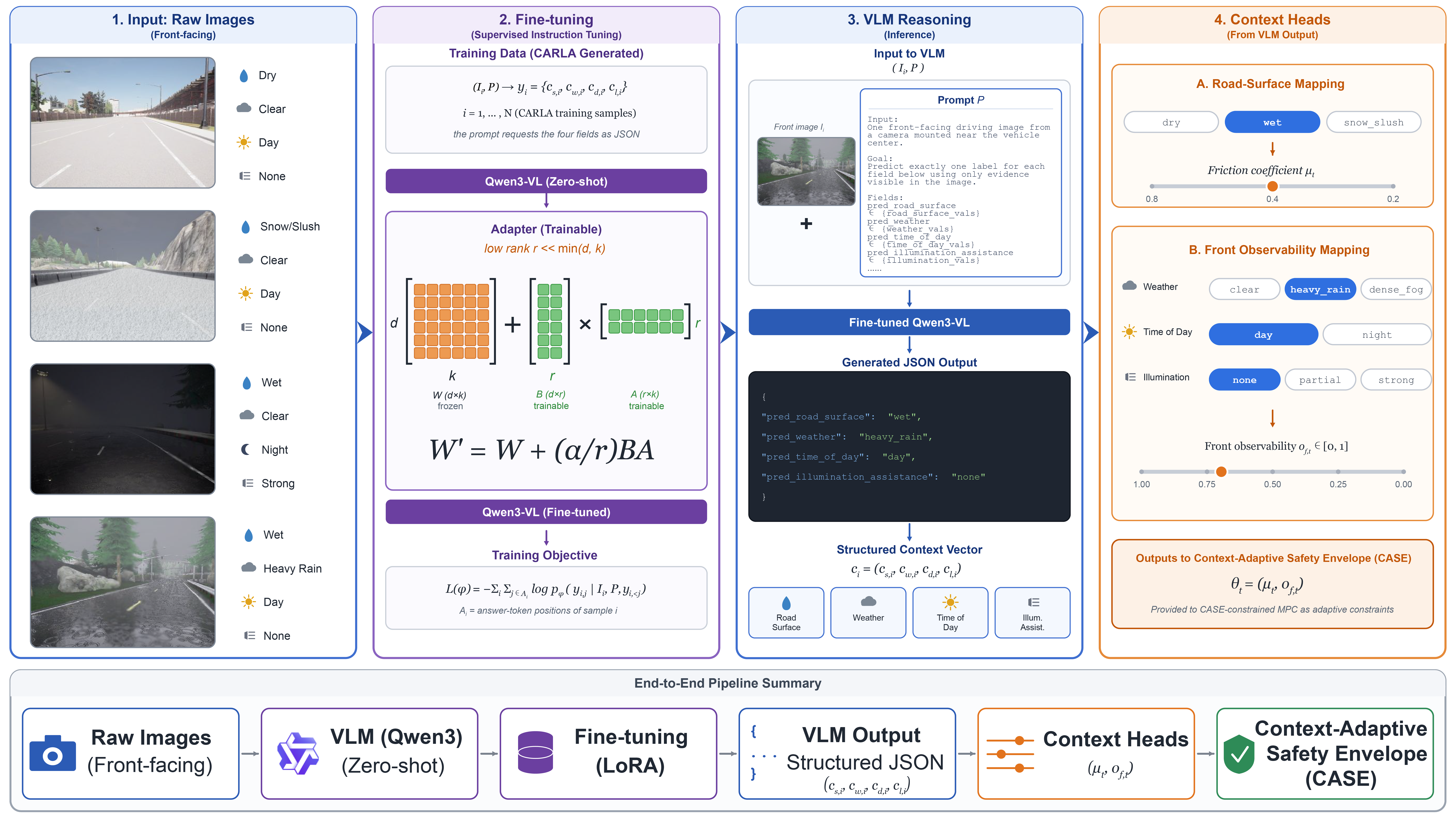}
  \caption{Overview of the VLM context inference pipeline.}
  \label{fig:vlm-pipeline}
\end{figure}

The VLM context inference module forms the upper level of the framework: it reasons about the driving scene from the front-camera image and produces the parameters that the safety envelope consumes. We present its parts in sequence: the scene reasoning the VLM performs and the prompt that drives it, the supervised fine-tuning that adapts the VLM to the task, and the mapping from the predicted context to the envelope parameters. Figure~\ref{fig:vlm-pipeline} summarizes the pipeline, from training images to the parameters consumed by the safety envelope.

\subsubsection{VLM reasoning}

The two parameters the safety envelope needs, the \emph{friction coefficient} and the \emph{forward observability}, are each governed by a few scene factors that a camera can resolve. The road surface governs the available tire--road friction, the vehicle's physical capability. Weather, time of day, and illumination assistance together govern how far ahead the scene can be reliably perceived, its forward observability. We therefore have the VLM reason about the scene through these four factors and report a discrete context
\begin{equation}
\mathbf{c} = \left[c_{s},\, c_{w},\, c_{d},\, c_{l}\right] \in \mathcal{C},
\label{eq:discrete-context}
\end{equation}
with road surface $c_{s} \in \{\text{dry}, \text{wet}, \text{snow}\}$, weather $c_{w} \in \{\text{clear}, \text{heavy rain}, \text{dense fog}\}$, time of day $c_{d} \in \{\text{day}, \text{night}\}$, and illumination assistance $c_{l} \in \{\text{none}, \text{partial}, \text{strong}\}$. Each label set is drawn from the available conditions in the CARLA simulator used for training and evaluation (the snow surface is overlaid manually, as CARLA has no built-in snow), so every category names a scene the framework actually encounters. Each field is a categorical judgment, which suits the nature of a vision-language model: built on a language backbone, it reasons reliably about discrete, semantic categories but is less reliable at predicting precise numerical values. The VLM therefore reports labels rather than numbers, and the mapping $\Phi$ converts them into the envelope parameters.

The reasoning is elicited by an instruction prompt $P$ that frames the task as scene-context classification. The prompt encodes a structured decision procedure, organized in five parts (Figure~\ref{fig:prompt}): (i) a \emph{role} that frames the model as a visual classifier of road-scene context; (ii) appearance-grounded \emph{label definitions} that give the model concrete criteria for each class, for example \emph{snow} as a surface with visible snow or slush cover; (iii) a \emph{decision order} that judges the surface from the pavement, the weather from the atmosphere and precipitation, the time of day from the ambient light, and the illumination assistance from the artificial light supporting forward perception; (iv) \emph{tie-break rules} for ambiguous scenes, such as assigning \emph{none} illumination assistance in daytime scenes; and (v) a strict \emph{output schema}, a single JSON object with the four fields and no explanation. Each response is checked to be valid JSON with every field holding an allowed label, so what the mapping receives is exactly the structured context it expects.

\begin{figure}[ht]
\centering
\fbox{\begin{minipage}{0.9\columnwidth}\small
You are a visual classifier for road-scene context.\\
Predict exactly one label per field from visible evidence.\\[4pt]
Fields:\\
\hspace*{1.5em}road\_surface $\in$ \{dry, wet, snow\}\\
\hspace*{1.5em}weather $\in$ \{clear, heavy\_rain, dense\_fog\}\\
\hspace*{1.5em}time\_of\_day $\in$ \{day, night\}\\
\hspace*{1.5em}illumination\_assistance $\in$ \{none, partial, strong\}\\[4pt]
Label definitions: appearance criteria for each label.\\
Decision order: surface, weather, time of day, illumination.\\
Tie-breaks: wet road, no rain $\to$ clear; day $\to$ none.\\
Output: one JSON object with the four fields; no explanation.
\end{minipage}}
\caption{Structure of the context-classification prompt $P$.}
\label{fig:prompt}
\end{figure}

\subsubsection{Fine-tuning with LoRA}

A compact open-source VLM is adapted to this task by supervised fine-tuning. For training sample $i$ with image $I_i$ and ground-truth context $\mathbf{c}_i$, the target $y_i$ is the context written as the JSON object the prompt requires, and the model learns to generate it token by token. The objective is the negative log-likelihood of the answer tokens, conditioned on the image and prompt,
\begin{equation}
\mathcal{L}(\phi) \;=\; -\sum_{i}\,\sum_{j \in \mathcal{A}_i} \log p_{\phi}\!\left(y_{i,j} \,\middle|\, I_i,\, P,\, y_{i,<j}\right),
\label{eq:nll}
\end{equation}
where $\mathcal{A}_i$ indexes the answer tokens and the prompt tokens are excluded from the loss, so training rewards the model for producing the correct labels.

Updating all of the VLM's weights would be costly and counterproductive here: the model has billions of parameters, and tuning them on a limited labeled set risks overfitting and eroding the broad world knowledge the framework depends on. Researchers have observed, however, that the weight change which specializes a large pretrained model to a narrow task tends to have low intrinsic rank \citep{huLoraLowrankAdaptation2022}. This motivates low-rank adaptation (LoRA), which freezes each pretrained weight matrix $W$ and learns a small additive update,
\begin{equation}
W' \;=\; W + \tfrac{\alpha}{r}\, B A, \qquad B \in \mathbb{R}^{d \times r},\; A \in \mathbb{R}^{r \times k},\; r \ll \min(d, k),
\label{eq:lora}
\end{equation}
with rank $r$ and scaling factor $\alpha$. The update $\tfrac{\alpha}{r}BA$, the trainable \emph{adapter}, is a product of two thin matrices, so it has rank at most $r$ and adds only $r(d+k)$ trainable parameters per matrix, against the $dk$ in $W$. The adapter thus recovers most of the benefit of full fine-tuning at a small fraction of the trainable parameters, leaving intact the base model's broad knowledge, on which the framework relies. The dataset and training configuration are detailed in Section~\ref{sec:experiments}.

\subsubsection{Context-to-parameter mapping}

The mapping $\Phi$ of Eq.~\eqref{eq:context} turns the categorical context into the two scalar parameters.

Friction is a table lookup on the road surface, $\mu = \Phi_{\mu}(c_{s})$, assigning each class a single coefficient with $\mu_{\mathrm{dry}} > \mu_{\mathrm{wet}} > \mu_{\mathrm{snow}}$. The three values are calibrated in CARLA so that each matches the friction the vehicle experiences on that surface.

Forward observability is a composite visibility index over the three perception-related fields. Each field value carries a visibility score in $[0,1]$, denoted $s_w(c_w)$, $s_d(c_d)$, and $s_l(c_l)$, with clear weather, daytime, and strong lighting scoring highest and the degraded values scoring lower. The index is the equal-weight mean of the active scores,
\begin{equation}
o_f \;=\; \frac{1}{|\mathcal{S}|}\sum_{s \in \mathcal{S}} s, \qquad
\mathcal{S} =
\begin{cases}
\{\, s_w(c_w),\, s_d(c_d) \,\}, & c_d = \text{day},\\[2pt]
\{\, s_w(c_w),\, s_d(c_d),\, s_l(c_l) \,\}, & c_d = \text{night},
\end{cases}
\label{eq:obs-map}
\end{equation}
clamped to $[0,1]$. Illumination assistance enters the average at night; in daylight it carries no usable visibility information and is dropped. The most degraded inputs, dense fog or heavy rain, night, and absent lighting, drive $o_f$ toward zero.

\subsection{Context-adaptive safety envelope}
\label{sec:rss-layer}

The envelope is defined on the vehicle's lane-frame state $\mathbf{x} = [e_y, e_\psi, v, s, \Delta v]$, comprising the lateral offset $e_y$ and heading error $e_\psi$ relative to the lane centerline, the speed $v$, the bumper-to-bumper gap $s$ to the lead vehicle, and the relative speed $\Delta v$. The control is $\mathbf{u} = [a, \delta]$, the longitudinal acceleration and front steering angle. The safety envelope is the intersection of three constraint sets,
\begin{equation}
\mathcal{U}_{\mathrm{safe}}(\mathbf{x},\boldsymbol{\theta}) \;=\; \mathcal{U}_{\mathrm{fric}}(\mathbf{x};\mu)\;\cap\;\mathcal{U}_{\mathrm{lon}}(\mathbf{x};\mu,o_f)\;\cap\;\mathcal{U}_{\mathrm{lat}}(\mathbf{x};\mu),
\label{eq:envelope}
\end{equation}
derived below.

\subsubsection{Physical capability: the shared friction budget}

The tire--road contact can transmit a bounded total force. In the vehicle body frame, the longitudinal acceleration is $a_x = a$ and the lateral (centripetal) acceleration is $a_y = v^2 \tan(\delta)/L$ for wheelbase $L$. With $g$ the gravitational acceleration, the friction circle requires
\begin{equation}
\sqrt{a_x^2 + a_y^2} \;\le\; \mu g .
\label{eq:friction-circle}
\end{equation}
Constraint~\eqref{eq:friction-circle} couples steering and braking through a single budget: lateral demand consumes braking authority and vice versa. It defines the friction-limited action set
\begin{equation}
\mathcal{U}_{\mathrm{fric}}(\mathbf{x};\mu) \;=\; \bigl\{\, \mathbf{u} = [a,\delta] : \sqrt{a_x^2 + a_y^2} \le \mu g \,\bigr\}.
\label{eq:u-fric}
\end{equation}

\subsubsection{Longitudinal safety: visibility-aware following distance}

Longitudinal safety follows the RSS proper-response argument with context-dependent parameters. If the lead vehicle brakes at up to $b_L$, while the ego accelerates at up to $\bar{a}$ during a response time $\rho_{\mathrm{lon}}$ before braking at $b(\mu) = \min\{\mu g,\, b_{\max}\}$, with $b_{\max}$ the mechanical braking limit, the minimum safe gap is
\begin{equation}
d_{\mathrm{lon}}(\mathbf{x};\boldsymbol{\theta}) \;=\; v\rho_{\mathrm{lon}} + \tfrac{1}{2}\bar{a}\rho_{\mathrm{lon}}^2 + \frac{\left(v + \bar{a}\rho_{\mathrm{lon}}\right)^2}{2\,b(\mu)} - \frac{v_L^2}{2\,b_L} \;+\; \Delta_{\mathrm{lon}}(v, o_f),
\label{eq:lon-distance}
\end{equation}
where $v$ and $v_L$ are the ego and lead speeds. The braking term $b(\mu)$ ties the ego's assumed stopping capability to the reported surface condition, so the safety limit reflects the vehicle's actual physical capability. 

The final term $\Delta_{\mathrm{lon}}(v, o_f)$ is an observability margin. Degraded perception cannot change the physics of stopping, but it widens the uncertainty about where hazards are, which the envelope absorbs as extra following distance. The margin grows with speed and shrinks as forward observability improves,
\begin{equation}
\Delta_{\mathrm{lon}}(v, o_f) \;=\; \left(d_0 + \tau v\right)\left(1 - o_f\right)^{\alpha}, \qquad \alpha \in (0,1],
\label{eq:lon-margin}
\end{equation}
where $d_0$ and $\tau$ set a base and a speed-proportional buffer, and the exponent $\alpha$ reflects the nonlinear relationship between observability and the margin: moderate observability loss produces only a small margin increase, while severe loss approaches the full buffer. The margin vanishes under ideal observability, recovering nominal RSS. Longitudinal safety then requires the gap to cover this distance, which defines
\begin{equation}
\mathcal{U}_{\mathrm{lon}}(\mathbf{x};\mu,o_f) \;=\; \bigl\{\, \mathbf{u} : s \ge d_{\mathrm{lon}}(\mathbf{x};\boldsymbol{\theta}) \,\bigr\}.
\label{eq:u-lon}
\end{equation}

\subsubsection{Lateral safety: an RSS rule for lane keeping}

Original RSS formulates lateral safety between two moving vehicles. Here, we re-derive it against the lane boundaries to ensure lane keeping. Let $v_{\mathrm{out}}^{L} = \max\{0,\, v\sin(e_\psi)\}$ and $v_{\mathrm{out}}^{R} = \max\{0,\, -v\sin(e_\psi)\}$ be the lateral speeds toward the left and right boundaries. In the worst case such a speed persists during a lateral response time $\rho_{\mathrm{lat}}$ and is then arrested at the lateral deceleration available under the friction budget, $a_y^{\mathrm{resp}} = \sqrt{(\mu g)^2 - a_x^2}$. The resulting worst-case drift toward each boundary is
\begin{equation}
d_{\mathrm{lat}}^{\sigma}(\mathbf{x},\mathbf{u};\mu) \;=\; v_{\mathrm{out}}^{\sigma}\,\rho_{\mathrm{lat}} \;+\; \frac{\left(v_{\mathrm{out}}^{\sigma}\right)^2}{2\,a_y^{\mathrm{resp}}}, \qquad \sigma \in \{L, R\}.
\label{eq:lat-distance}
\end{equation}
Consider a vehicle of length $L_v$ and width $w_v$ in a lane of width $w$. When the vehicle yaws relative to the lane, its corners swing outward, so it reaches further toward the boundaries than its width alone: the distance from its center to its outer edge, measured across the lane, is $\tfrac{1}{2}\!\left(L_v|\sin e_\psi| + w_v|\cos e_\psi|\right)$ and grows with the heading error $e_\psi$. With $e_y$ the lateral offset from the centerline (positive toward the right), the clearances to the two boundaries are
\begin{equation}
\begin{aligned}
c_L &= \tfrac{1}{2}w + e_y - \tfrac{1}{2}\!\left(L_v|\sin e_\psi| + w_v|\cos e_\psi|\right), \\
c_R &= \tfrac{1}{2}w - e_y - \tfrac{1}{2}\!\left(L_v|\sin e_\psi| + w_v|\cos e_\psi|\right).
\end{aligned}
\label{eq:lat-clearance}
\end{equation}
Lateral safety requires each drift to stay within its clearance, which defines
\begin{equation}
\mathcal{U}_{\mathrm{lat}}(\mathbf{x};\mu) \;=\; \bigl\{\, \mathbf{u} : c_L \ge d_{\mathrm{lat}}^{L},\; c_R \ge d_{\mathrm{lat}}^{R} \,\bigr\}.
\label{eq:u-lat}
\end{equation} The drift depends on the control through $a_x$: braking harder while drifting outward reduces $a_y^{\mathrm{resp}}$ and lengthens the drift, a coupling absent from decoupled formulations.

\subsection{VLM-CASE-MPC: an MPC realization}
\label{sec:mpc-layer}

Any controller whose actions stay inside $\mathcal{U}_{\mathrm{safe}}$ satisfies the safety constraints, independent of how it is realized. We realize the framework with MPC and refer to the resulting controller as VLM-CASE-MPC.

\subsubsection{State and prediction model}

Within one solve at control time $t$, the controller predicts the state $\mathbf{x}$ of Section~\ref{sec:rss-layer} over steps $k = 0, \dots, N$, written $\mathbf{x}_k = [e_{y,k}, e_{\psi,k}, v_k, s_k, \Delta v_k]$ and initialized from the observed state. The lead state sets $s_0$ and $\Delta v_0$ when a valid lead is present. Otherwise, $s_0$ takes a large default value and $\Delta v_0 = 0$, which deactivates car following. The prediction model is a kinematic bicycle in the lane frame,
\begin{equation}
\dot{e}_y = v\sin(e_\psi), \quad
\dot{e}_\psi = \frac{v}{L}\tan(\delta) - v\cos(e_\psi)\,\kappa_{\mathrm{ref}}, \quad
\dot{v} = a, \quad
\dot{s} = \Delta v, \quad
\dot{\Delta v} = -a,
\label{eq:dynamics}
\end{equation}
with $\kappa_{\mathrm{ref}}$ the reference-path curvature and a constant-speed lead assumption over the horizon, discretized by forward Euler with interval $\Delta t$ to give $\mathbf{x}_{k+1} = f(\mathbf{x}_k, \mathbf{u}_k)$.

\subsubsection{Objective}

The stage cost serves both driving tasks together with control smoothness:
\begin{equation}
\begin{aligned}
\ell(\mathbf{x}_k, \mathbf{u}_k) \;=\;{}& w_y\, e_{y,k}^2 + w_\psi\, e_{\psi,k}^2 + w_v \left(v_k - v_{\mathrm{ref}}\right)^2 + \beta\, w_s \left(s_k - s^{\mathrm{des}}_k\right)^2 \\
{}&+ w_a\, a_k^2 + w_\delta\, \delta_k^2 + w_{\Delta a} \left(a_k - a_{k-1}\right)^2 + w_{\Delta\delta} \left(\delta_k - \delta_{k-1}\right)^2.
\end{aligned}
\label{eq:stage-cost}
\end{equation}
The lane-keeping terms $w_y\, e_{y,k}^2$ and $w_\psi\, e_{\psi,k}^2$ penalize the lateral offset and heading error from the lane centerline. The car-following terms $w_v (v_k - v_{\mathrm{ref}})^2$ and $\beta\, w_s (s_k - s^{\mathrm{des}}_k)^2$ track the reference speed and, when a lead is present, the desired spacing $s^{\mathrm{des}}_k = d_h + \tau_h v_k$, a standstill gap $d_h$ plus a time-headway term. The remaining terms penalize control effort and its rate of change. Here $v_{\mathrm{ref}}$ is the lead speed when a lead is present and the desired cruising speed otherwise, $\beta \in \{0,1\}$ is the lead-valid flag, and the weights $w_y, \dots, w_{\Delta\delta}$ are nonnegative. The terminal cost $\ell_N$ applies the same lane-keeping and car-following terms at the final state $\mathbf{x}_N$.

On a curve, the centripetal demand $v^2|\kappa|$ consumes friction that braking may suddenly need. To anticipate this, the objective adds a cornering penalty
\begin{equation}
J_{\mathrm{corner}} \;=\; w_c \sum_{k=0}^{N} \max\{0,\; v_k^2\,|\kappa_k| - \eta\,\mu_t\, g \}^{2},
\label{eq:cornering}
\end{equation}
where $w_c \ge 0$ weights the penalty and $\eta \in (0,1)$ reserves a fraction of the friction budget for cornering. Because $\kappa_k$ is the reference curvature ahead of the vehicle, the penalty acts before the curve is reached. When the VLM reports a low-friction surface, $\mu_t$ shrinks, the reserved budget tightens, and the controller reduces speed in advance, mimicking a human driver who eases off before a snowy curve.

\subsubsection{Complete problem}

At each control time $t$, the controller solves
\begin{equation}
\begin{aligned}
\min_{\mathbf{u}_{0:N-1}} \quad & \sum_{k=0}^{N-1} \ell\left(\mathbf{x}_k, \mathbf{u}_k\right) + \ell_N\left(\mathbf{x}_N\right) + J_{\mathrm{corner}} \\
\text{s.t.} \quad & \mathbf{x}_{k+1} = f(\mathbf{x}_k, \mathbf{u}_k), \\
& \mathbf{x}_0 = \mathbf{x}_t, \\
& \mathbf{u}_k \in \mathcal{U}_{\mathrm{safe}}(\mathbf{x}_k, \boldsymbol{\theta}_t),
\end{aligned}
\label{eq:mpc}
\end{equation}
together with actuator bounds and rate limits. The safety envelope is imposed at every step of the horizon with the current context-derived parameters $\boldsymbol{\theta}_t$. The controller thus optimizes performance while respecting the safety constraints that the VLM's scene understanding implies. The first optimized action is applied to the vehicle, and the problem is re-solved at the next step in receding-horizon fashion.

\section{Experimental setup}
\label{sec:experiments}

We evaluate the framework in closed-loop simulation with the CARLA simulator on two maps that cover complementary operating regimes: Town03, a dense urban map driven at 40--60~km/h, and Town04, which includes high-speed highway segments driven at 70--90~km/h. The ego vehicle performs combined lane keeping and car following: it tracks the lane centerline while regulating its speed and, when a lead vehicle is present, the following distance. The VLM updates the context every 2~s while the controller runs at 20~Hz, so the two operate asynchronously. The evaluation comprises three experiment groups organized around the two envelope parameters: two groups vary one factor each, the tire--road friction $\mu$ or the forward observability $o_f$, and the third degrades both at once as an integrated stress test. Together the groups span 66 scenarios, each defined by a map, a condition, and a speed, and repeated over three runs, for 198 evaluation runs in total. The rest of this section details the scenarios, the controllers compared, the VLM inference settings, and the evaluation metrics.

\subsection{Driving scenarios}
\label{sec:scenarios}

\begin{figure}[!htb]
  \centering
  \includegraphics[width=\textwidth]{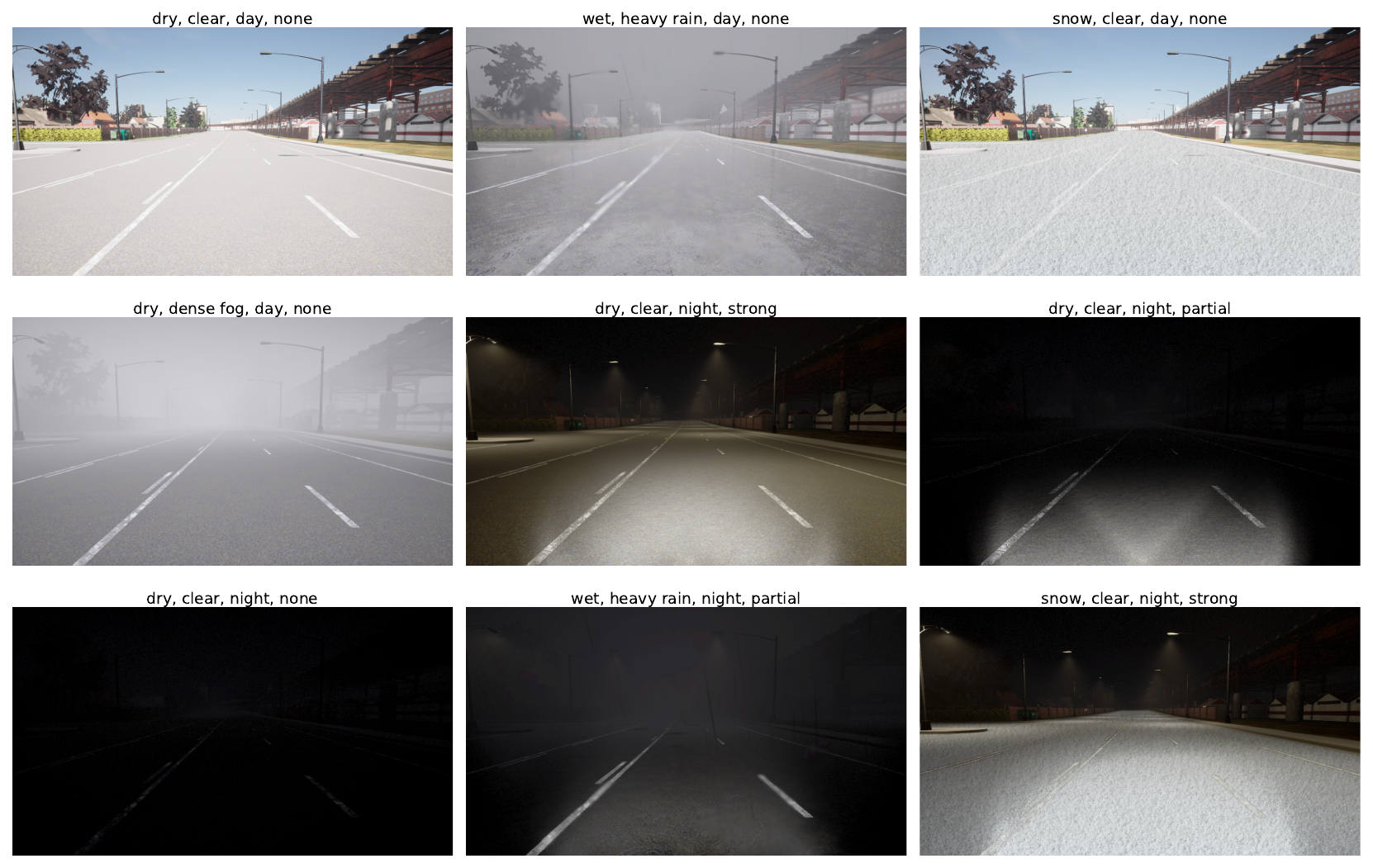}
  \caption{Front-camera frames for the evaluated conditions (road surface, weather, time of day, illumination).}
  \label{fig:gallery}
\end{figure}

The ego drives a 350~m route on Town03 and a 1000~m route on Town04, each mixing curved and straight segments, and each scenario is swept over three speeds per map, 40, 50, and 60~km/h on Town03 and 70, 80, and 90~km/h on Town04. The swept speed is the ego's target speed in the no-lead group and the lead vehicle's speed in the two following groups. CARLA provides no native snow surface, so snow scenes overlay a snow and slush texture on the road with the tire friction reduced accordingly. Figure~\ref{fig:gallery} shows representative front-camera frames across the conditions. The three experimental groups are:

\begin{itemize}
  \item \textbf{No-lead driving} (18 scenarios: 3 surfaces $\times$ 2 maps $\times$ 3 speeds). Tests lateral safety under friction alone. The ego drives without a lead vehicle on dry (N1, $\mu = 0.8$), wet (N2, $\mu = 0.4$), and snow (N3, $\mu = 0.2$) surfaces.
  \item \textbf{Constant-lead following} (30 scenarios: 5 conditions $\times$ 2 maps $\times$ 3 speeds). Tests the following distance under visibility alone. The ego follows a constant-speed lead in clear day (L1), foggy day (L2), and clear night with strong (L3), partial (L4), and no (L5) illumination assistance.
  \item \textbf{Lead braking} (18 scenarios: 3 conditions $\times$ 2 maps $\times$ 3 speeds). The integrated stress test with both factors degraded. The lead travels at constant speed and then brakes hard to a stop at 40\% of the route, under three compound conditions: C1, dry surface in clear daylight; C2, wet surface in heavy rain at night with partial illumination; and C3, snow surface on a clear night with strong illumination.
\end{itemize}

\FloatBarrier

\subsection{Controllers}
\label{sec:controllers}

To validate the effectiveness of the proposed VLM-CASE-MPC, we compare it against three controllers: Base MPC, VLM-MPC, and Fixed-Envelope MPC. Their settings are detailed as follows:

\begin{itemize}
  \item \textbf{Base MPC} is the vanilla MPC controller with no context-adaptive safety envelope, while all other settings remain the same. It has both lane-keeping and car-following abilities.
  \item \textbf{VLM-MPC} reproduces the recent VLM controller of \citet{longVLMMPCModelPredictive2026}. It relies on the VLM to reason about the driving conditions and adjust MPC parameters such as the desired speed, headway, and cost weights accordingly. Because the original controller is longitudinal only, we supplement it with a basic waypoint-following lateral controller so that it can keep the lane in our combined task.
  \item \textbf{Fixed-Envelope MPC} applies the full safety envelope but freezes the context at nominal good-condition values ($\mu = 0.8$, $o_f = 1$). Therefore, its margins never adapt to the conditions the vehicle meets.
  \item \textbf{VLM-CASE-MPC} is the proposed controller, in which the live VLM reasoning continuously drives the envelope parameters $\boldsymbol{\theta} = [\mu, o_f]$, tightening the margins as conditions worsen and relaxing them as conditions improve.
\end{itemize}

To attribute the effect of each adaptation, we additionally evaluate two variants of the proposed controller: adapting friction only and adapting visibility only, each freezing the other parameter at its nominal value.

\subsection{VLM inference settings}
\label{sec:vlm-setup}

The backbone model used in this research is Qwen3-VL, a state-of-the-art open-source VLM. Among its variants we adopt the 8B model: as the accuracy results below show, it is the strongest on illumination assistance, the field that most separates the variants, while keeping an inference time compatible with the asynchronous control loop. We further fine-tune it with LoRA to ensure reliable performance on our driving scenes. The fine-tuning dataset comprises 10{,}560 front-camera frames at $1280 \times 720$, collected in CARLA across the road surfaces, weather types, times of day, and illumination settings of Section~\ref{sec:vlm-layer} and labeled with the four context fields, split into 8{,}448 training, 1{,}056 validation, and 1{,}056 test frames stratified over the joint distribution of the four fields. The adapters use rank $r = 16$ and scaling factor $\alpha = 32$, applied to the query and value projections of the last twelve transformer layers. Training runs for 2{,}000 optimization steps at a learning rate of $1 \times 10^{-4}$ with an effective batch size of 8 (a per-step batch of 1 accumulated over 8 steps), about two passes over the training set, and the checkpoint with the highest validation accuracy is retained.

Table~\ref{tab:vlm} reports per-field and overall accuracy on the held-out test set. Among the zero-shot base models the 2B model is by far the weakest, at 8.6\% all-fields accuracy. The 4B and 8B models perform comparably on road surface, weather, and time of day, but they diverge on illumination assistance: accuracy rises from 30.3\% (2B) to 68.9\% (4B) to 76.6\% (8B). The 8B model leads on this field by a clear margin, at a zero-shot inference time of 1.30~s that still fits the asynchronous update budget. However, none of the base models reliably gets all four fields right at once, the 8B reaching only 42.5\% all-fields accuracy, so we fine-tune with LoRA. Adaptation raises every field above 99\% and the all-fields accuracy to 98.3\%, and the adapted model infers in 1.18~s.

\begin{table}[!htbp]
  \centering
  \footnotesize
  \caption{Scene-context accuracy (\%) and inference time: LoRA versus zero-shot base models.}
  \label{tab:vlm}
  \resizebox{\textwidth}{!}{
\begin{tabular}{llcccccc}
\toprule
Model & Variant & Road surface & Weather & Time of day & Illumination & Overall & Inference (s) \\
\midrule
Qwen3-VL-2B & Zero-shot & 78.88 & 44.41 & 99.15 & 30.30 & 8.62 & 0.69 \\
Qwen3-VL-4B & Zero-shot & 84.75 & 57.39 & 98.58 & 68.94 & 43.75 & 0.91 \\
Qwen3-VL-8B & Zero-shot & 84.47 & 56.72 & 98.30 & 76.61 & 42.52 & 1.30 \\
Qwen3-VL-8B & LoRA fine-tuned & 99.43 & 99.81 & 100.00 & 99.05 & 98.30 & 1.18 \\
\bottomrule
\end{tabular}
}
\end{table}

\subsection{Evaluation metrics}
\label{sec:metrics}

We assess each controller on driving safety and lane-keeping quality. A run is a success if the vehicle completes the route within its lane or, in the lead-braking group, stops safely behind the lead. Departing the lane or stopping inside the minimum safe gap counts as a failure. From the runs we compute the following metrics.

\begin{itemize}
  \item \textbf{Success rate} is the fraction of runs that succeed. It is the primary measure of safe task completion: a controller with a higher success rate more reliably keeps the vehicle in its lane and avoids collision across the route.
  \item \textbf{Failure rate per kilometer} is the number of failures divided by the total distance driven, expressed per kilometer. Normalizing by distance makes the measure comparable across routes of different length and across runs that terminate early. A lower value is safer.
  \item \textbf{Minimum time-to-collision (TTC)} is a widely used surrogate safety metric that measures the time that would remain before two road users collide if they maintained their current motion. A larger TTC indicates a safer interaction. In our setting the relevant interaction is car following, so we evaluate the TTC between the ego and the lead vehicle. For each run we record the minimum TTC, and we report the average of these minimum values across all runs. Because TTC captures the longitudinal margin to the lead, we aggregate it only over runs without a lane-keeping failure, since a vehicle that has left the lane no longer faces the lead and its TTC is no longer meaningful.
  \item \textbf{Minimum boundary clearance} is the lateral margin left at the most critical moment of a run: the distance between the vehicle side and the nearer lane edge, $c = \tfrac{1}{2}(w - w_v) - |e_y|$, where $w$ is the lane width and $w_v$ the vehicle width. For each run we record the minimum clearance, and we report the average of these minima across runs. A positive per-run minimum means the vehicle kept a margin to the lane boundary throughout the run; a negative one means the boundary was crossed.
  \item \textbf{Mean lateral error} is the absolute deviation of the vehicle from the lane centerline, averaged over each run and then over runs. It reflects typical tracking accuracy; smaller values indicate tighter lane keeping.
\end{itemize}

\section{Results}
\label{sec:results}

We present the lead-braking group first, the main safety test with both friction and visibility degraded at once, followed by the two single-factor groups, no-lead under varying friction and constant-lead under varying visibility, and an ablation study that separates the two adaptations.

\subsection{Lead braking experiments under combined adverse conditions}
\label{sec:res-leadbrake}

Table~\ref{tab:leadbrake} reports the lead-braking results. Over the three conditions, the Base MPC succeeds in 51.9\% of its runs, the VLM-MPC in 75.9\%, and the Fixed-Envelope MPC in 81.5\%, while the VLM-CASE-MPC completes all 54 runs. The controllers diverge as the conditions worsen. On the dry condition (C1) all of them succeed. On the wet condition (C2) the Base MPC drops to 50.0\% and the VLM-MPC to 77.8\%. On the snow condition (C3) the Base MPC reaches only 5.6\%, and the VLM-MPC and Fixed-Envelope MPC both fall to 50.0\%. The VLM-CASE-MPC stays at 100\% on both adverse conditions.

\begin{figure}[!htbp]
  \centering
  \includegraphics[width=0.85\textwidth]{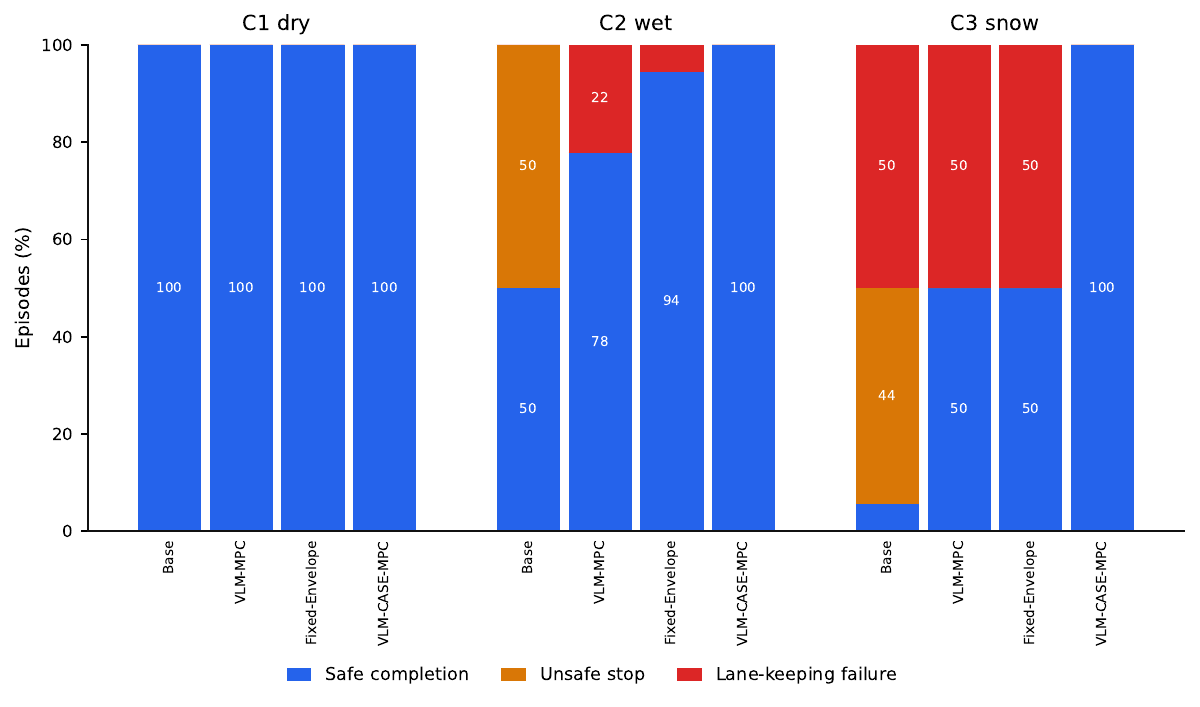}
  \caption{Run outcomes in lead-braking experiments by condition and controller.}
  \label{fig:leadbrake-outcomes}
\end{figure}

\begin{figure}[!htbp]
  \centering
  \includegraphics[width=0.95\textwidth]{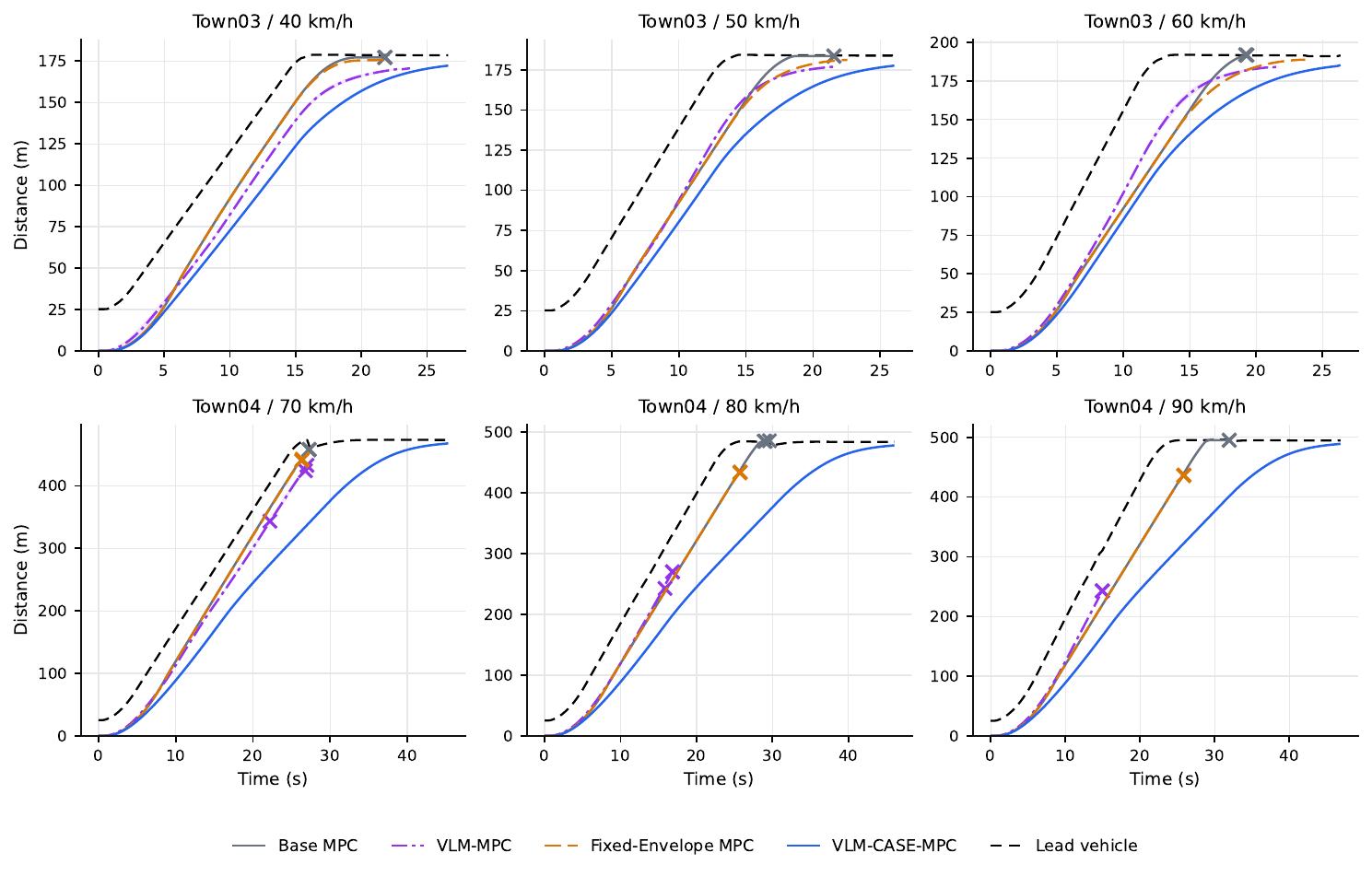}
  \caption{Trajectories under the snow condition (C3) in lead-braking experiments.}
  \label{fig:leadbrake-trace}
\end{figure}

The three baselines fail for different reasons, as the outcome breakdown in Figure~\ref{fig:leadbrake-outcomes} shows. The Base MPC has no safety guarantee, so it fails in every way. It keeps no following margin: its minimum TTC drops to 0.96~s on wet and 0.27~s on snow, against 4.76~s and 4.72~s for the VLM-CASE-MPC, so it nearly contacts the stopped lead. It also has no friction-aware control, so on snow it slides out of the lane. Its snow failures are thus a mix of unsafe stops and lane departures. The Fixed-Envelope MPC keeps a fixed following gap. In these scenarios the gap is large enough to stop for the sudden brake, so the car-following rarely fails. Its problem is the frozen friction: assuming dry grip on snow, it brakes and steers harder than the surface allows and leaves the lane, its minimum boundary clearance falling to 0.08~m on average across runs. The VLM-MPC follows the lead safely. It keeps a minimum TTC above 3~s to avoid any rear-end conflict. Its failures stem from the lack of a tire--road friction model: the controller does not account for the reduced grip on wet and snow, and keeps a speed the surface cannot support on curves. On a curve the available grip is then too small to hold the vehicle, which runs off the lane or slides out, giving four heading-error failures on wet and nine lane departures on snow. This is especially apparent in the high-speed, low-friction runs in Figure~\ref{fig:leadbrake-trace}.

The VLM-CASE-MPC avoids all three failure modes. Told that grip is low, it lowers the assumed braking capability $b(\mu)$, couples braking and steering through the shared friction budget, and widens the following gap in advance. It therefore brakes within the budget and stays in the lane, keeping a minimum boundary clearance of 0.71~m on snow, where the failing baselines retain 0.16~m or less.

\begin{table}[!htbp]
  \centering
  \caption{Lead-braking experimental results by condition and controller. $\uparrow$ means higher is better, $\downarrow$ means lower is better; the best result is in bold.}
  \label{tab:leadbrake}
  \resizebox{\textwidth}{!}{
\begin{tabular}{llrrrrr}
\toprule
Condition & Controller & \multicolumn{1}{c}{Success (\%)} & \multicolumn{1}{c}{Failure/km $\downarrow$} & \multicolumn{1}{c}{Min TTC (s) $\uparrow$} & \multicolumn{1}{c}{Min clearance (m) $\uparrow$} & \multicolumn{1}{c}{Mean $|e_y|$ (m) $\downarrow$} \\
\midrule
C1 dry & Base MPC & 100.00 & 0.00 & 1.51 $\pm$ 0.42 & \textbf{0.764 $\pm$ 0.022} & 0.0074 $\pm$ 0.0041 \\
 & VLM-MPC & 100.00 & 0.00 & \textbf{3.48 $\pm$ 0.04} & n/a & n/a \\
 & Fixed-Envelope MPC & 100.00 & 0.00 & 2.72 $\pm$ 0.21 & 0.761 $\pm$ 0.021 & \textbf{0.0070 $\pm$ 0.0040} \\
 & \textbf{VLM-CASE-MPC (Ours)} & 100.00 & 0.00 & 2.73 $\pm$ 0.20 & 0.762 $\pm$ 0.020 & \textbf{0.0070 $\pm$ 0.0040} \\
\midrule
C2 wet & Base MPC & 50.00 & 1.50 & 0.96 $\pm$ 0.89 & 0.732 $\pm$ 0.045 & 0.0155 $\pm$ 0.0126 \\
 & VLM-MPC & 77.78 & 0.70 & 4.18 $\pm$ 0.24 & n/a & n/a \\
 & Fixed-Envelope MPC & 94.44 & 0.17 & 2.71 $\pm$ 0.18 & 0.595 $\pm$ 0.209 & 0.0196 $\pm$ 0.0166 \\
 & \textbf{VLM-CASE-MPC (Ours)} & \textbf{100.00} & \textbf{0.00} & \textbf{4.76 $\pm$ 0.17} & \textbf{0.759 $\pm$ 0.030} & \textbf{0.0083 $\pm$ 0.0071} \\
\midrule
C3 snow & Base MPC & 5.56 & 2.85 & 0.27 $\pm$ 0.38 & 0.155 $\pm$ 0.532 & 0.0583 $\pm$ 0.0313 \\
 & VLM-MPC & 50.00 & 2.09 & 3.42 $\pm$ 0.30 & n/a & n/a \\
 & Fixed-Envelope MPC & 50.00 & 1.62 & 2.39 $\pm$ 0.57 & 0.079 $\pm$ 0.683 & 0.0451 $\pm$ 0.0399 \\
 & \textbf{VLM-CASE-MPC (Ours)} & \textbf{100.00} & \textbf{0.00} & \textbf{4.72 $\pm$ 0.13} & \textbf{0.712 $\pm$ 0.071} & \textbf{0.0300 $\pm$ 0.0281} \\
\bottomrule
\end{tabular}
}
  \par\smallskip
  \begin{minipage}{\linewidth}
    \footnotesize \textit{Note:} Min TTC is over runs without a lane-keeping failure; other metrics are over all runs. Lateral metrics are omitted for VLM-MPC (no lateral control).
  \end{minipage}
\end{table}

\FloatBarrier
\subsection{No-lead driving experiments under varying road friction}
\label{sec:res-friction}

Table~\ref{tab:nolead} reports the no-lead results. On the dry and wet surfaces all controllers complete every run and keep the lane almost identically, with mean lateral error near 0.02--0.03~m and minimum boundary clearance of at least 0.65~m. The performance diverges on snow. The Base and Fixed-Envelope controllers complete only 33\% of the snow runs (6 of 18): they succeed at the lowest speed on each map but leave the lane at every higher speed. The VLM-CASE-MPC completes all 18.

\begin{table}[!htbp]
  \centering
  \caption{No-lead experimental results by road surface and controller. $\downarrow$ means lower is better; the best result is in bold.}
  \label{tab:nolead}
  \resizebox{\textwidth}{!}{
\begin{tabular}{llrrrrr}
\toprule
Condition & Controller & \multicolumn{1}{c}{Success (\%)} & \multicolumn{1}{c}{Failure/km $\downarrow$} & \multicolumn{1}{c}{Mean speed (km/h)} & \multicolumn{1}{c}{Min clearance (m) $\uparrow$} & \multicolumn{1}{c}{Mean $|e_y|$ (m) $\downarrow$} \\
\midrule
N1 dry & Base MPC & 100.00 & 0.00 & 58.8 $\pm$ 15.8 & \textbf{0.719 $\pm$ 0.040} & \textbf{0.0198 $\pm$ 0.0112} \\
 & Fixed-Envelope MPC & 100.00 & 0.00 & 58.8 $\pm$ 15.8 & 0.718 $\pm$ 0.042 & \textbf{0.0198 $\pm$ 0.0113} \\
 & \textbf{VLM-CASE-MPC (Ours)} & 100.00 & 0.00 & 58.8 $\pm$ 15.8 & \textbf{0.719 $\pm$ 0.042} & 0.0199 $\pm$ 0.0112 \\
\midrule
N2 wet & Base MPC & 100.00 & 0.00 & 58.1 $\pm$ 15.6 & \textbf{0.699 $\pm$ 0.047} & \textbf{0.0264 $\pm$ 0.0180} \\
 & Fixed-Envelope MPC & 100.00 & 0.00 & 58.1 $\pm$ 15.6 & 0.698 $\pm$ 0.050 & \textbf{0.0264 $\pm$ 0.0182} \\
 & \textbf{VLM-CASE-MPC (Ours)} & 100.00 & 0.00 & 57.9 $\pm$ 15.5 & 0.650 $\pm$ 0.150 & 0.0316 $\pm$ 0.0306 \\
\midrule
N3 snow & Base MPC & 33.33 & 1.67 & 48.7 $\pm$ 13.6 & -0.073 $\pm$ 0.614 & 0.0565 $\pm$ 0.0270 \\
 & Fixed-Envelope MPC & 33.33 & 1.68 & 48.7 $\pm$ 13.6 & -0.140 $\pm$ 0.636 & \textbf{0.0546 $\pm$ 0.0239} \\
 & \textbf{VLM-CASE-MPC (Ours)} & \textbf{100.00} & \textbf{0.00} & 44.0 $\pm$ \hphantom{0}5.5 & \textbf{0.184 $\pm$ 0.214} & 0.1966 $\pm$ 0.1258 \\
\bottomrule
\end{tabular}
}
  \par\smallskip
  \begin{minipage}{\linewidth}
    \footnotesize \textit{Note:} Mean speed is over successful runs; other metrics are over all runs. VLM-MPC is omitted (no lateral control).
  \end{minipage}
\end{table}

\begin{figure}[!htbp]
  \centering
  \includegraphics[width=0.7\textwidth]{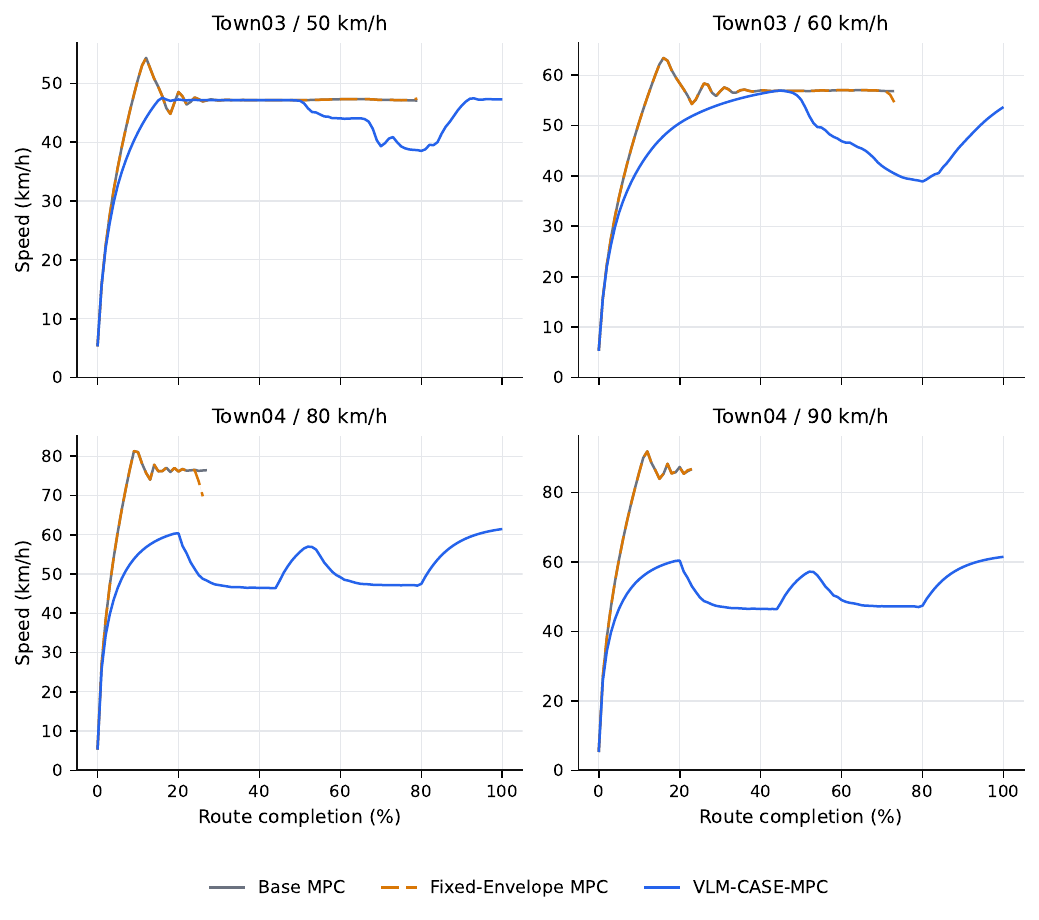}
  \caption{Ego speed profiles in no-lead experiments on the snow surface.}
  \label{fig:nolead-snow}
\end{figure}

The failure is friction-limited cornering. On a curve, the centripetal demand $v^2|\kappa|$ must fit within the available grip $\mu g$. On snow ($\mu = 0.2$) this budget is small. A controller that carries too high a speed into the curve exceeds this budget and slides out of the lane, which is what the Base and Fixed-Envelope controllers do at 50~km/h and above. The VLM-CASE-MPC is told that the surface is snow and lowers $\mu$. A lower $\mu$ tightens both the friction constraint of Eq.~\eqref{eq:friction-circle} and the cornering penalty of Eq.~\eqref{eq:cornering}, so the controller slows before each curve, to a mean of 44~km/h on snow. Figure~\ref{fig:nolead-snow} shows the effect in the speed profiles: the non-adaptive controllers hold their speed until they depart, while the proposed controller slows through each curve and completes the route, as human drivers do. At the cost of a modest reduction in speed, the controller eliminates loss-of-control failures entirely.

The two lateral columns in Table~\ref{tab:nolead} should be read together. On snow the Base and Fixed-Envelope controllers show a smaller mean lateral error (about $0.05$~m) than the VLM-CASE-MPC ($0.20$~m), a counting effect: a failed run ends the moment the vehicle leaves the lane, so the failing controllers are scored mostly on the easy driving before their departure, while the VLM-CASE-MPC is scored on the full route at every speed. The boundary clearance resolves the comparison: every run of the VLM-CASE-MPC ends with a positive minimum clearance, averaging $0.18$~m, while the runs that crossed the lane edge pull the failing controllers' averages below zero ($-0.07$ and $-0.14$~m).

\FloatBarrier
\subsection{Constant-lead following experiments under varying visibility}
\label{sec:res-visibility}

Constant-speed following poses no emergency, and all four controllers complete every run in this group (Table~\ref{tab:constlead}). They also keep the lane equally well, with mean lateral error near 0.02~m and minimum boundary clearance near 0.72~m in every condition (VLM-MPC excepted, as it steers with a separate tracker). What sets the controllers apart is the following gap, the room left if the lead behaves unexpectedly: it stays near 24~m for the Base and Fixed-Envelope controllers and near 51~m for VLM-MPC, whereas for the VLM-CASE-MPC it widens as visibility falls. We therefore examine this gap in detail.

\begin{figure}[!htbp]
  \centering
  \includegraphics[width=\textwidth]{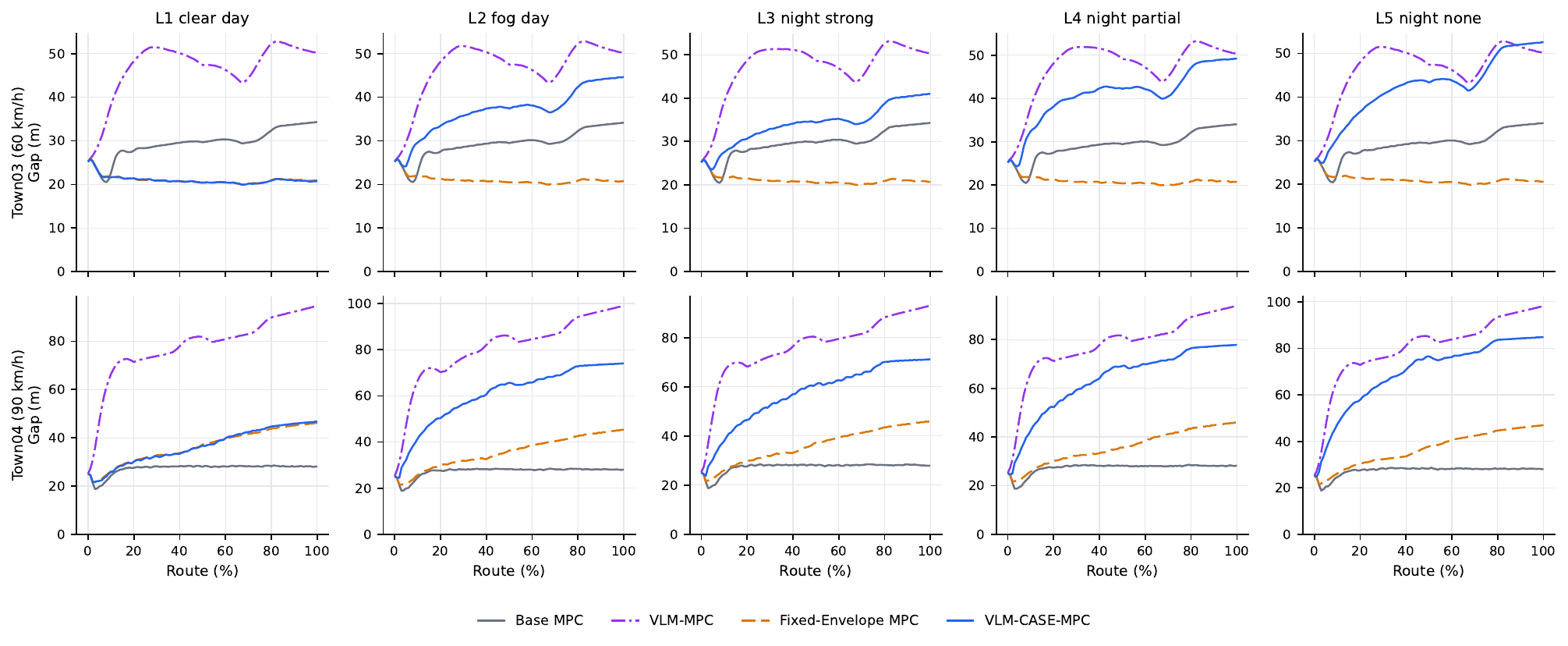}
  \caption{Following gap versus route completion under varying visibility.}
  \label{fig:constlead-gap}
\end{figure}

Figure~\ref{fig:constlead-gap} shows the gap as visibility falls from clear day (L1) to dark night (L5), for both maps at their top speed (Town03 at 60~km/h and Town04 at 90~km/h). The Base and Fixed-Envelope controllers keep the same gap pattern in every condition, around 27~m and 35~m, and do not respond to visibility. The VLM-CASE-MPC widens its gap as the forward observability drops, through the visibility margin of Eq.~\eqref{eq:lon-margin}: at 90~km/h on Town04 its mean gap at top speed opens from 36~m in clear conditions to 66~m on the darkest night. In clear conditions the margin vanishes, so its gap coincides with the Fixed-Envelope baseline. The adaptation engages only as perception degrades. The VLM-MPC also varies its gap with the conditions, but only slightly, between about 72 and 76~m. This small variation is not what keeps it safe. Its safety comes from a conservative following preference that holds a large gap even in clear and moderate visibility, around 73~m, and does not open much further as the night darkens. The two controllers are therefore safe in different ways: VLM-MPC by staying conservative in every condition, and the VLM-CASE-MPC by adapting, keeping a moderate gap when sight is clear and opening it up as the night deepens, the way a careful human driver does.

\begin{table}[!htbp]
  \centering
  \caption{Constant-lead experimental results by visibility condition and controller. $\downarrow$ means lower is better; the best result is in bold.}
  \label{tab:constlead}
  \resizebox{\textwidth}{!}{
\begin{tabular}{llrrrr}
\toprule
Condition & Controller & \multicolumn{1}{c}{Success (\%)} & \multicolumn{1}{c}{Mean gap (m)} & \multicolumn{1}{c}{Min clearance (m) $\uparrow$} & \multicolumn{1}{c}{Mean $|e_y|$ (m) $\downarrow$} \\
\midrule
L1 clear day & Base MPC & 100.00 & 23.8 $\pm$ \hphantom{0}4.2 & 0.719 $\pm$ 0.029 & 0.0195 $\pm$ 0.0112 \\
 & VLM-MPC & 100.00 & 50.7 $\pm$ 13.2 & n/a & n/a \\
 & Fixed-Envelope MPC & 100.00 & 23.9 $\pm$ \hphantom{0}6.3 & 0.723 $\pm$ 0.035 & 0.0188 $\pm$ 0.0107 \\
 & \textbf{VLM-CASE-MPC (Ours)} & 100.00 & 24.0 $\pm$ \hphantom{0}6.3 & \textbf{0.725 $\pm$ 0.032} & \textbf{0.0187 $\pm$ 0.0106} \\
\midrule
L2 fog day & Base MPC & 100.00 & 23.7 $\pm$ \hphantom{0}4.2 & 0.717 $\pm$ 0.032 & 0.0194 $\pm$ 0.0114 \\
 & VLM-MPC & 100.00 & 51.3 $\pm$ 13.8 & n/a & n/a \\
 & Fixed-Envelope MPC & 100.00 & 23.8 $\pm$ \hphantom{0}6.1 & \textbf{0.722 $\pm$ 0.035} & 0.0189 $\pm$ 0.0108 \\
 & \textbf{VLM-CASE-MPC (Ours)} & 100.00 & 41.3 $\pm$ 10.3 & 0.718 $\pm$ 0.030 & \textbf{0.0184 $\pm$ 0.0094} \\
\midrule
L3 night strong & Base MPC & 100.00 & 23.8 $\pm$ \hphantom{0}4.2 & 0.717 $\pm$ 0.032 & 0.0197 $\pm$ 0.0113 \\
 & VLM-MPC & 100.00 & 50.6 $\pm$ 12.7 & n/a & n/a \\
 & Fixed-Envelope MPC & 100.00 & 24.2 $\pm$ \hphantom{0}6.1 & \textbf{0.719 $\pm$ 0.034} & 0.0187 $\pm$ 0.0107 \\
 & \textbf{VLM-CASE-MPC (Ours)} & 100.00 & 38.7 $\pm$ \hphantom{0}9.9 & \textbf{0.719 $\pm$ 0.028} & \textbf{0.0186 $\pm$ 0.0095} \\
\midrule
L4 night partial & Base MPC & 100.00 & 23.8 $\pm$ \hphantom{0}4.2 & 0.719 $\pm$ 0.030 & 0.0195 $\pm$ 0.0111 \\
 & VLM-MPC & 100.00 & 51.2 $\pm$ 12.8 & n/a & n/a \\
 & Fixed-Envelope MPC & 100.00 & 23.9 $\pm$ \hphantom{0}6.1 & \textbf{0.722 $\pm$ 0.033} & 0.0190 $\pm$ 0.0106 \\
 & \textbf{VLM-CASE-MPC (Ours)} & 100.00 & 44.2 $\pm$ 10.3 & \textbf{0.722 $\pm$ 0.028} & \textbf{0.0180 $\pm$ 0.0095} \\
\midrule
L5 night none & Base MPC & 100.00 & 23.8 $\pm$ \hphantom{0}4.1 & 0.717 $\pm$ 0.031 & 0.0195 $\pm$ 0.0113 \\
 & VLM-MPC & 100.00 & 51.4 $\pm$ 13.7 & n/a & n/a \\
 & Fixed-Envelope MPC & 100.00 & 24.0 $\pm$ \hphantom{0}6.4 & \textbf{0.720 $\pm$ 0.033} & 0.0189 $\pm$ 0.0106 \\
 & \textbf{VLM-CASE-MPC (Ours)} & 100.00 & 47.6 $\pm$ 12.0 & 0.719 $\pm$ 0.028 & \textbf{0.0184 $\pm$ 0.0088} \\
\bottomrule
\end{tabular}
}
  \par\smallskip
  \begin{minipage}{\linewidth}
    \footnotesize \textit{Note:} Mean gap is over successful runs; other metrics are over all runs. Lateral metrics are omitted for VLM-MPC (no lateral control).
  \end{minipage}
\end{table}

\FloatBarrier
\subsection{Ablation study}
\label{sec:res-ablation}

Table~\ref{tab:ablation} isolates the two adaptations on the lead-braking task, each variant adapting one parameter and freezing the other at its nominal value. The three conditions stress them differently. The dry condition (C1) stresses neither, with good grip and good visibility, and every variant succeeds. Each adverse condition stresses one of them: the snow condition (C3) is friction-limited, with low grip ($\mu = 0.2$) but fair visibility ($o_f = 0.83$), while the wet condition (C2) is visibility-limited, with moderate grip ($\mu = 0.4$) but a rainy night and only partial illumination, the poorest visibility of the three ($o_f = 0.5$). In both adverse conditions the envelope keeps the vehicle from rear-ending the lead, so the failures that remain are lateral, and each adaptation removes the failures in the condition that matches it.

On snow, friction adaptation keeps the vehicle in the lane. Friction-only clears the condition completely (100\% success) and keeps a minimum boundary clearance of $0.71$~m, because the correct grip lets the controller brake and steer within the snow friction budget. The variants that do not adapt friction, visibility-only and the Fixed-Envelope baseline, assume more grip than the surface offers, steer harder than it allows, and slide off the lane: their minimum clearance collapses to about $0.1$~m on average and only half of their runs succeed.

On wet, the remaining failure is again lateral, but visibility adaptation prevents it through the longitudinal margin. Visibility-only widens the following gap to match the low observability, raising the minimum TTC to $4.56$~s, against $2.93$~s for friction-only. This is the coupling that matters: the larger margin lets the controller brake gently when the lead stops, so it spends little of the friction budget on braking and keeps enough for steering, holding its heading and clearing every run. Friction-only, keeping a tighter gap, must brake hard, exhausts the shared budget, and suffers a heading-error failure.

Neither adaptation covers the other condition: friction-only stays at 94\% on wet and visibility-only at 50\% on snow, matching the non-adaptive Fixed-Envelope baseline. Only the full controller, which adapts both, reaches 100\% on both. The two adaptations are therefore complementary, each covering the condition the other cannot.

\begin{table}[!htbp]
  \centering
  \caption{Ablation of the proposed controller on the lead-braking task. $\uparrow$ means higher is better, $\downarrow$ means lower is better; the best result is in bold.}
  \label{tab:ablation}
  \resizebox{\textwidth}{!}{
\begin{tabular}{llrrrrr}
\toprule
Condition & Controller & \multicolumn{1}{c}{Success (\%)} & \multicolumn{1}{c}{Failure/km $\downarrow$} & \multicolumn{1}{c}{Min TTC (s) $\uparrow$} & \multicolumn{1}{c}{Min clearance (m) $\uparrow$} & \multicolumn{1}{c}{Mean $|e_y|$ (m) $\downarrow$} \\
\midrule
C1 dry & Fixed-Envelope MPC & 100.00 & 0.00 & 2.72 $\pm$ 0.21 & 0.761 $\pm$ 0.021 & 0.0070 $\pm$ 0.0040 \\
 & Visibility-only adaptation & 100.00 & 0.00 & \textbf{2.74 $\pm$ 0.17} & 0.761 $\pm$ 0.021 & 0.0070 $\pm$ 0.0040 \\
 & Friction-only adaptation & 100.00 & 0.00 & 2.73 $\pm$ 0.18 & 0.761 $\pm$ 0.022 & \textbf{0.0069 $\pm$ 0.0041} \\
 & \textbf{VLM-CASE-MPC (Ours)} & 100.00 & 0.00 & 2.73 $\pm$ 0.20 & \textbf{0.762 $\pm$ 0.020} & 0.0070 $\pm$ 0.0040 \\
\midrule
C2 wet & Fixed-Envelope MPC & 94.44 & 0.17 & 2.71 $\pm$ 0.18 & 0.595 $\pm$ 0.209 & 0.0196 $\pm$ 0.0166 \\
 & Visibility-only adaptation & \textbf{100.00} & \textbf{0.00} & 4.56 $\pm$ 0.14 & 0.711 $\pm$ 0.122 & 0.0100 $\pm$ 0.0084 \\
 & Friction-only adaptation & 94.44 & 0.17 & 2.93 $\pm$ 0.11 & 0.723 $\pm$ 0.066 & 0.0112 $\pm$ 0.0072 \\
 & \textbf{VLM-CASE-MPC (Ours)} & \textbf{100.00} & \textbf{0.00} & \textbf{4.76 $\pm$ 0.17} & \textbf{0.759 $\pm$ 0.030} & \textbf{0.0083 $\pm$ 0.0071} \\
\midrule
C3 snow & Fixed-Envelope MPC & 50.00 & 1.62 & 2.39 $\pm$ 0.57 & 0.079 $\pm$ 0.683 & 0.0451 $\pm$ 0.0399 \\
 & Visibility-only adaptation & 50.00 & 1.67 & 3.53 $\pm$ 0.40 & 0.108 $\pm$ 0.666 & 0.0419 $\pm$ 0.0387 \\
 & Friction-only adaptation & \textbf{100.00} & \textbf{0.00} & 3.56 $\pm$ 0.12 & 0.708 $\pm$ 0.068 & 0.0332 $\pm$ 0.0296 \\
 & \textbf{VLM-CASE-MPC (Ours)} & \textbf{100.00} & \textbf{0.00} & \textbf{4.72 $\pm$ 0.13} & \textbf{0.712 $\pm$ 0.071} & \textbf{0.0300 $\pm$ 0.0281} \\
\bottomrule
\end{tabular}
}
  \par\smallskip
  \begin{minipage}{\linewidth}
    \footnotesize \textit{Note:} Visibility-only and friction-only each adapt a single parameter, with the other frozen at its nominal value. Min TTC is over runs without a lane-keeping failure; other metrics are over all runs.
  \end{minipage}
\end{table}

\section{Conclusion and discussion}
\label{sec:conclusion}

This paper introduced a framework that gives an autonomous vehicle anticipatory, human-like caution under adverse driving conditions: a VLM interprets the scene and translates its environmental understanding into the vehicle's safety limits. The model reads the road surface and visibility from a front camera and parametrizes a context-adaptive safety envelope that couples braking and steering through a shared friction budget and widens the following margin as perception degrades; an MPC controller then drives freely within the envelope. Because the VLM sets the envelope's parameters, this caution stays bounded by a physically grounded safety model at all times, and because the VLM runs asynchronously, it does not slow the real-time control loop.

Closed-loop experiments in CARLA support the design. On the no-lead snow surface, friction adaptation eliminated loss-of-control failures, lifting the success rate from 33\% to 100\% by constraining the vehicle to a moderate cornering speed within the friction budget. In constant-speed following, visibility adaptation widened the mean following gap, averaged over all runs, from 24~m in clear daylight (L1) to 48~m on the darkest night (L5), while the non-adaptive controllers held a fixed gap. On the integrated emergency-braking task, the proposed controller completed all 54 trials (100\%), against 82\% for Fixed-Envelope MPC, 76\% for VLM-MPC, and 52\% for Base MPC. The ablation confirmed that the two adaptations are complementary: adapting friction alone solved the friction-limited condition, where the correct grip is what keeps the vehicle in its lane; adapting visibility alone solved the visibility-limited condition, where a wider following margin is what averts the emergency; only the full controller, supplying both, succeeded in both conditions.

Several limitations point to future work. The evaluation is in simulation, and transferring the framework to real vehicles will require closing the appearance gap and validating the friction calibration on real surfaces. The context is a small set of discrete categories converted to envelope parameters by a calibrated table; a continuous mapping from richer scene descriptions could capture finer gradations of condition and would let the envelope respond to situations outside the present category set. Finally, the safety envelope is controller-agnostic, and pairing it with controllers other than MPC, or composing it with formal runtime monitors, is a natural extension of this work.

\section*{CRediT authorship contribution statement}
\textbf{Tianjia Yang:} Writing -- review \& editing, Writing -- original draft, Methodology, Software, Investigation, Conceptualization. \textbf{Ke Li:} Writing -- review \& editing, Validation, Investigation. \textbf{Ruwen Qin:} Writing -- review \& editing, Conceptualization. \textbf{Xianbiao Hu:} Writing -- review \& editing, Supervision, Methodology, Conceptualization.

\bibliographystyle{elsarticle-harv}
\bibliography{vlm-rss-mpc}

\end{document}